\documentclass{article}
\usepackage{spconf,amsmath,graphicx,hyperref}
\usepackage{amssymb,algorithm,algorithmic,booktabs,multirow,adjustbox,parskip}

\title{SHEFL: Sparse Heterogeneous Ensemble Federated Learning with Group-Balanced Aggregation}
\twoauthors
 {Keumseo Ryum, Joonhyuk Kang\sthanks{\textit{(Corresponding author: Joonhyuk Kang)}. This work was partly supported by the Institute of Information \& Communications Technology Planning \& Evaluation (IITP)-ITRC (Information Technology Research Center) grant funded by the Korea government (MSIT) (IITP-2026-RS-2020-II201787, contribution rate: 50\%) and (IITP-2026-RS-2023-00259991, contribution rate: 25\%).}}
	{KAIST\\
	School of Electrical Engineering\\
	Daejeon, Republic of Korea}
 {Jinu Gong\sthanks{This research was financially supported by Hansung University for Jinu Gong (contribution rate: 25\%).}}
	{Hansung University\\
	Department of Applied Artificial Intelligence\\
	Seoul, Republic of Korea}

\begin{document}
%
\maketitle

\begin{abstract}
System heterogeneity in federated learning (FL) requires training workloads and communication demands to be adapted to client resource capacities. Federated ensemble learning distributes multiple predictors across clients, but assigning one ensemble member to each client reduces the number of contributors to each member as the ensemble grows. We propose SHEFL, a global ensemble-based FL framework that uses the additional computation of high-power clients (HPCs) to train all ensemble members while low-power clients (LPCs) train one assigned member. SHEFL addresses increased HPC uplink demand and biased aggregation toward HPC updates through resource-aware sparse update allocation and group-balanced aggregation. Across four vision datasets, SHEFL improves over Fed-Ensemble (FedEns) in all data heterogeneity settings. Under a matched total uplink budget, SHEFL reaches the target accuracy in fewer rounds than FedEns and improves final accuracy by up to 14.65 points. Ablations further show that group-balanced aggregation improves accuracy and convergence under different resource allocations.
\end{abstract}

\begin{keywords}
Federated learning, System heterogeneity, Deep ensembles, Top-$k$ sparsification
\end{keywords}

\section{Introduction}

Federated learning (FL) \cite{mcmahan2017communication} has emerged as a distributed learning framework in which multiple edge devices collaboratively train a global model without sharing their local data, thereby preserving data privacy and avoiding the transfer of raw data to a central server. However, FL deployed in practical scenarios suffers from heterogeneity in computational resources, system resources, and communication capabilities among clients \cite{fedprox}. Such heterogeneity creates stragglers, i.e., devices with limited computational capabilities that require longer training times and consequently delay the overall training process. 
Handling system heterogeneity is therefore essential, as differences in client capabilities can significantly slow training and degrade convergence. 

Several approaches addressing system heterogeneity in FL have focused on model heterogeneity, since clients with limited resources cannot efficiently train large global models \cite{wu2024fedekt}. Model-heterogeneous FL  enables resource-constrained clients to train submodels of the global model \cite{heterofl, kim2023depthfl, fedrolex}, while distillation-based methods combine predictions from heterogeneous client models on the server \cite{feddf, fedbe, fedet}. However, aggregating models with different structures requires compatible parameter mappings, while distillation methods may require additional public or proxy data at the server. These approaches primarily accommodate resource-constrained clients, while additional computation at more capable clients can also be used to train multiple global predictors.
\begin{figure*}[!t]
\centering
\newlength{\figgap}
\setlength{\figgap}{0.004\textwidth}

\makebox[\textwidth][c]{%
\begin{minipage}[t]{0.249\textwidth}
    \centering
    \includegraphics[width=\linewidth]{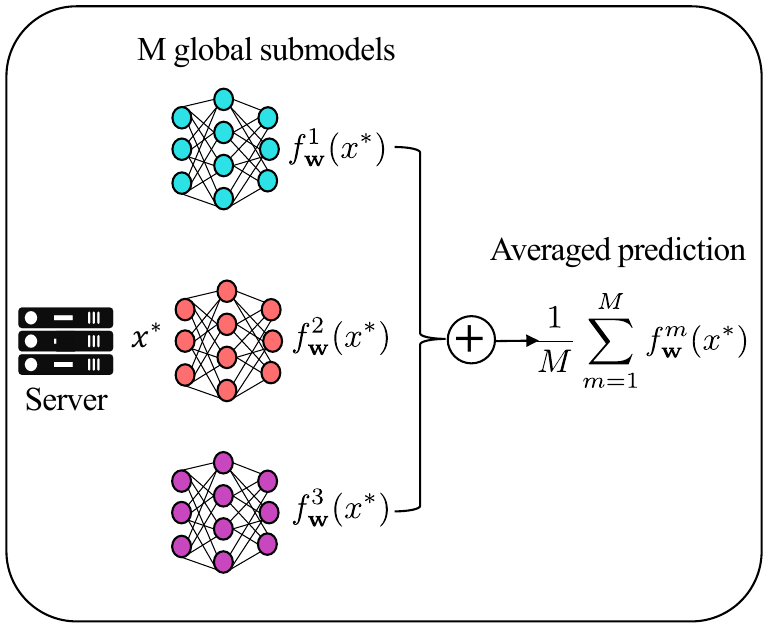}
    \vspace{0.5mm}
    {\footnotesize (a) Global ensemble prediction}
\end{minipage}%
\hspace{\figgap}%
\begin{minipage}[t]{0.299\textwidth}
    \centering
    \includegraphics[width=\linewidth]{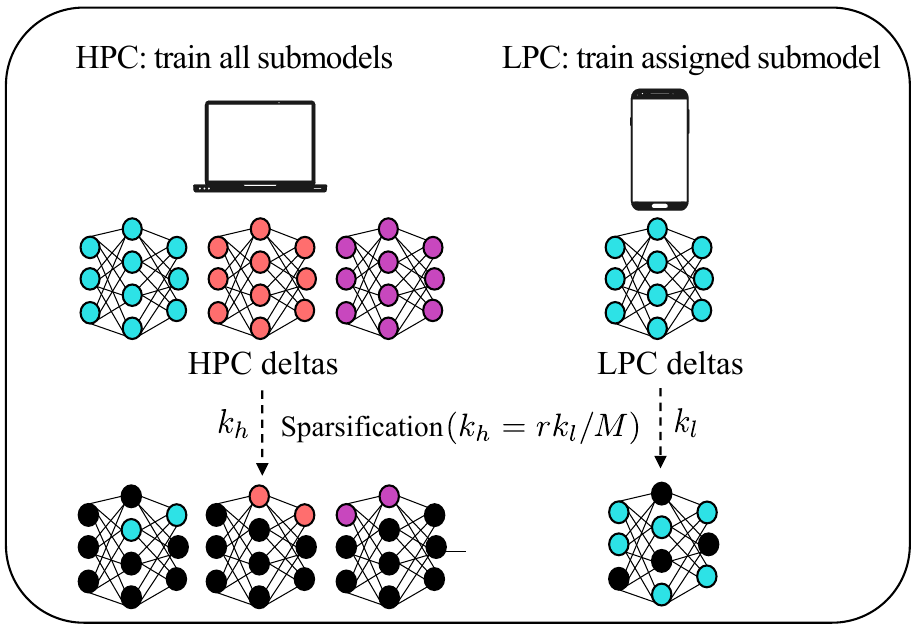}
    \vspace{0.5mm}
    {\footnotesize (b) Resource-aware sparse uplink}
\end{minipage}%
\hspace{\figgap}%
\begin{minipage}[t]{0.444\textwidth}
    \centering
    \includegraphics[width=\linewidth]{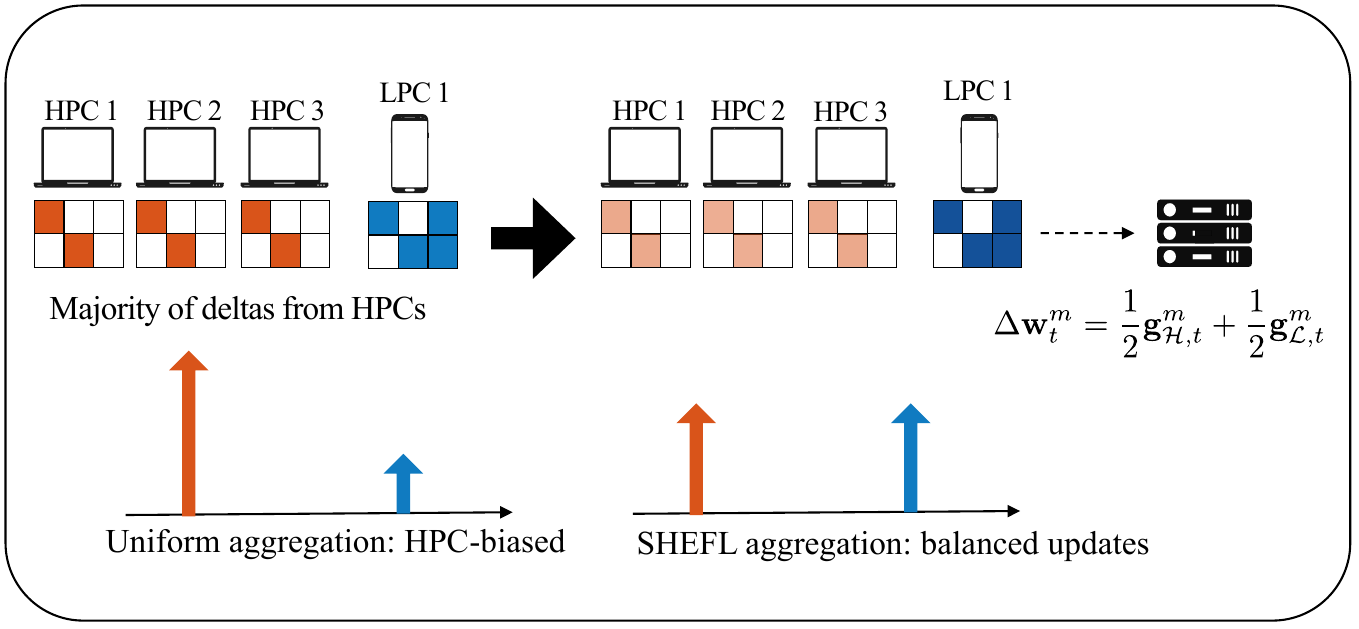}
    \vspace{0.5mm}
    {\footnotesize (c) Group-balanced aggregation}
\end{minipage}%
}

\caption{Overview of SHEFL. 
(a) The server maintains $M$ global submodels and averages their logits for ensemble prediction. 
(b) HPCs train all submodels while LPCs train one assigned submodel, and both groups upload top-$k$ sparse deltas under group-specific budgets. 
(c) The server mitigates HPC-biased updates using group-balanced aggregation.}
\label{fig:shefl_overview}
\end{figure*}

 Deep ensembles \cite{deepensembles} combine predictions from multiple models, providing a model structure across which client training workloads can be distributed in FL. Ensemble distillation methods \cite{feddf,fedbe} combine knowledge from multiple client models into a single global model under heterogeneous client data, but do not allocate ensemble training workloads according  to client computational capability. Fed-Ensemble (FedEns) \cite{shi2023fed} maintains $M$ global models and assigns each participating client to one ensemble member, so each client trains only one model per round. Each member consequently receives updates from only $C/M$ of $C$ participating clients on average. As $M$ increases, the reduced participation per member can delay ensemble training, and the uniform workload also leaves additional computation at capable clients unused.

Motivated by these observations, we introduce SHEFL, a heterogeneous ensemble FL framework in which high-power clients (HPCs) train all $M$ ensemble members while low-power clients (LPCs) train one assigned member. Our contributions are as follows: 
\begin{enumerate}
    \item We propose SHEFL, a heterogeneous ensemble FL framework that uses additional HPC computation to sustain participation in each ensemble member.
    \item We introduce resource-aware sparse update allocation and group-balanced aggregation to control uplink demand and contribution imbalance under asymmetric client workloads.
    \item Experiments across four vision datasets show the effectiveness of SHEFL under different data heterogeneity, ensemble sizes, and communication budgets. We further verify the effect of our proposed aggregation under different resource allocations via ablation studies.
\end{enumerate}
\section{SHEFL Framework}
\label{sec:shefl}
We consider a federated learning system with a server and \(N\) clients. The server maintains \(M\) global models \(\{w^m\}_{m=1}^{M}\), each with \(d\) parameters, and averages their logits for ensemble prediction. Clients are divided into high-power clients (HPCs) \(\mathcal{H}\) and low-power clients (LPCs) \(\mathcal{L}\) according to their computational capabilities. At round \(t\), the server selects participating clients \(\mathcal{N}_t=\mathcal{H}_t\cup\mathcal{L}_t\), where \(\mathcal{H}_t\subseteq\mathcal{H}\) and \(\mathcal{L}_t\subseteq\mathcal{L}\).

Each selected HPC trains all \(M\) models, while each selected LPC trains one model following the permutation-based assignment of FedEns [11]. Model assignments are shuffled over blocks of \(M\) rounds to distribute LPC training across the ensemble. We denote the HPCs and LPCs contributing to model \(m\) by \(\mathcal{H}_t^m\) and \(\mathcal{L}_t^m\), respectively, with \(\mathcal{H}_t^m=\mathcal{H}_t\). This assignment increases participation in each member but also changes the uplink demand and the composition of updates received by the server.

\subsection{Sparse Uplink Allocation}


To control the uplink demand induced by this assignment, each client transmits a sparse delta after local training. For client \(i\) in resource group \(g\in\{H,L\}\), the transmitted update for model \(m\) is
\begin{equation}
\Delta_{i,t}^m =
\mathrm{TopK}\!\left(
\mathbf{w}_{i,t}^m-\mathbf{w}_{t-1}^m, k_g
\right),
\qquad g\in\{H,L\}.
\label{eq:sparse_delta}
\end{equation}
Here, \(w_{i,t}^m\) is the locally trained model initialized from \(w_{t-1}^m\), and \(\operatorname{TopK}\) retains the \(k_g\) coordinates with the largest magnitudes. The budgets \(k_H\) and \(k_L\) apply to each model update from an HPC and an LPC, respectively.

Since an HPC uploads \(M\) updates and an LPC uploads one, the total uplink budget ratio between one LPC and one HPC is \(1:r\), where

\begin{equation}
    r=\frac{Mk_H}{k_L},
    \qquad
    k_H=\frac{rk_L}{M}.
\label{eq:resource_ratio}
\end{equation}
\begin{table*}[!t]
  \centering
  \caption{Final test accuracy (\%) under Dirichlet and pathological data heterogeneity averaged over five seeds.}
  \label{tab:acc_combined}
  \vspace{2pt}
  \setlength{\tabcolsep}{3pt}
  \renewcommand{\arraystretch}{1}
  \begin{adjustbox}{max width=\textwidth}
  \begin{tabular}{@{}l
                  cc   
                  cc   
                  cc   
                  cc@{}} 
    \toprule[1.2pt]
    \textbf{Method}
      & \multicolumn{2}{c}{\textbf{FMNIST}}
      & \multicolumn{2}{c}{\textbf{CIFAR10}}
      & \multicolumn{2}{c}{\textbf{CIFAR100}}
      & \multicolumn{2}{c}{\textbf{SVHN}} \\
    \cmidrule(lr){2-3}
    \cmidrule(lr){4-5}
    \cmidrule(lr){6-7}
    \cmidrule(lr){8-9}
      & $\mathrm{Dir}$(0.6) & Path(4)
      & $\mathrm{Dir}$(0.6) & Path(4)
      & $\mathrm{Dir}$(0.6) & Path(20)
      & $\mathrm{Dir}$(0.6) & Path(4) \\
    \midrule
    FedAvg
      & 89.78$\pm$0.28
      & 85.20$\pm$4.52
      & 76.94$\pm$1.02
      & 70.85$\pm$2.13
      & 63.15$\pm$0.34
      & 59.36$\pm$0.70
      & 93.62$\pm$0.11
      & 92.28$\pm$0.22 \\

    FedProx
      & 89.72$\pm$0.20
      & 86.96$\pm$2.59
      & 77.14$\pm$0.55
      & 71.76$\pm$1.78
      & 63.12$\pm$0.49
      & \textbf{59.43}$\pm$0.62
      & 93.43$\pm$0.11
      & {92.47}$\pm$0.08 \\

    FedNova
      & 89.63$\pm$0.38
      & \textbf{87.95}$\pm$0.40
      & 76.19$\pm$1.85
      & 60.98$\pm$4.78
      & \textbf{65.75}$\pm$0.49
      & 59.33$\pm$1.32
      & \textbf{94.38}$\pm$0.19
      & 89.82$\pm$2.04 \\
\midrule
    HeteroFL
      & 88.24$\pm$0.31
      & 83.60$\pm$1.45
      & 70.27$\pm$1.36
      & 51.48$\pm$5.33
      & 59.12$\pm$0.59
      & 51.39$\pm$0.97
      & 93.95$\pm$0.36
      & 80.64$\pm$8.14 \\
\midrule
    FedBE
      & 88.57$\pm$0.36
      & 81.62$\pm$5.77
      & 68.65$\pm$1.36
      & 55.40$\pm$9.60
      & 52.89$\pm$2.39
      & 48.13$\pm$1.61
      & 92.54$\pm$1.01
      & 82.22$\pm$5.33 \\

    FedEns
      & {89.92}$\pm$0.29
      & 83.14$\pm$3.10
      & 77.79$\pm$0.94
      & 70.25$\pm$2.56
      & 62.01$\pm$0.43
      & 55.64$\pm$0.87
      & 93.67$\pm$0.18
      & 91.51$\pm$0.30 \\

    SHEFL
      & \textbf{90.14}$\pm$0.20
      & 86.96$\pm$1.53
      & \textbf{78.63}$\pm$0.90
      & \textbf{72.74}$\pm$2.76
      & 63.46$\pm$0.48
      & 59.12$\pm$0.95
      & {93.86}$\pm$0.18
      & \textbf{92.54}$\pm$0.48 \\

    \bottomrule[1.2pt]
  \end{tabular}
  \end{adjustbox}
\end{table*}
The allocation ratio determines how the fixed per-round sparse update
budget is distributed between the two client groups.
Let $K$ denote the total number of retained coordinates across all
participating clients in one round.
For \(C_H=|\mathcal{H}_t|\) and \(C_L=|\mathcal{L}_t|\), a fixed total budget of \(K\) retained coordinates per round satisfies
\begin{equation}
    C_H M k_H + C_L k_L = K.
    \label{eq:total_budget}
\end{equation}
Combining Eqs.~\eqref{eq:resource_ratio} and
\eqref{eq:total_budget} gives
\begin{equation}
    k_L
    =
    \frac{K}{C_L+rC_H},
    \qquad
    k_H
    =
    \frac{rK}{M(C_L+rC_H)}.
    \label{eq:group_budget}
\end{equation}
The default allocation uses \(r=M\), giving \(k_H=k_L\). Under a fixed \(K\), decreasing \(r\) shifts the sparse uplink budget from HPC updates to LPC updates.

\subsection{Balanced Aggregation}
Sparse uplink allocation controls the number of transmitted coordinates, while aggregation determines how the received updates contribute to each model. Since all participating HPCs train every member and only the assigned LPCs contribute to it, the two groups can have different numbers of contributors. When both groups are present, their average updates are
\begin{equation}
\begin{aligned}
g_{\mathcal{H},t}^m
&=
\frac{1}{|\mathcal{H}_t^m|}
\sum_{i\in\mathcal{H}_t^m}
\Delta_{i,t}^m, \quad
g_{\mathcal{L},t}^m =
\frac{1}{|\mathcal{L}_t^m|}
\sum_{i\in\mathcal{L}_t^m}
\Delta_{i,t}^m.
\end{aligned}
\label{eq:group_updates}
\end{equation}
When both groups participate in model $m$, uniform aggregation can be
written as
\begin{equation}
\begin{aligned}
\frac{1}{|\mathcal{N}_t^m|}
\sum_{i\in\mathcal{N}_t^m}
\Delta_{i,t}^m
&=
\frac{|\mathcal{H}_t^m|}
     {|\mathcal{N}_t^m|}
g_{\mathcal{H},t}^m
+
\frac{|\mathcal{L}_t^m|}
     {|\mathcal{N}_t^m|}
g_{\mathcal{L},t}^m,
\end{aligned}
\label{eq:uniform_aggregation}
\end{equation}
where \(\mathcal{N}_t^m=\mathcal{H}_t^m\cup\mathcal{L}_t^m\). Uniform aggregation therefore assigns group weights in proportion to contributor counts. SHEFL uses group-balanced aggregation to give the two group averages equal weight whenever both groups contribute.

\begin{equation}
\Delta_t^m
=
\frac{1}{2}g_{\mathcal{H},t}^m
+
\frac{1}{2}g_{\mathcal{L},t}^m.
\label{eq:group_balanced_aggregation}
\end{equation}
When only one group contributes, its average update receives the full weight. If neither group contributes, the corresponding global model remains unchanged. The server applies the resulting delta to \(w_{t-1}^m\) to obtain \(w_t^m\).
This group normalization prevents contributor count alone
from determining each resource group's aggregation mass. An overview of our method is shown in Fig.~\ref{fig:shefl_overview}.

\begin{figure}[t]
\centering
\includegraphics[width=\columnwidth]{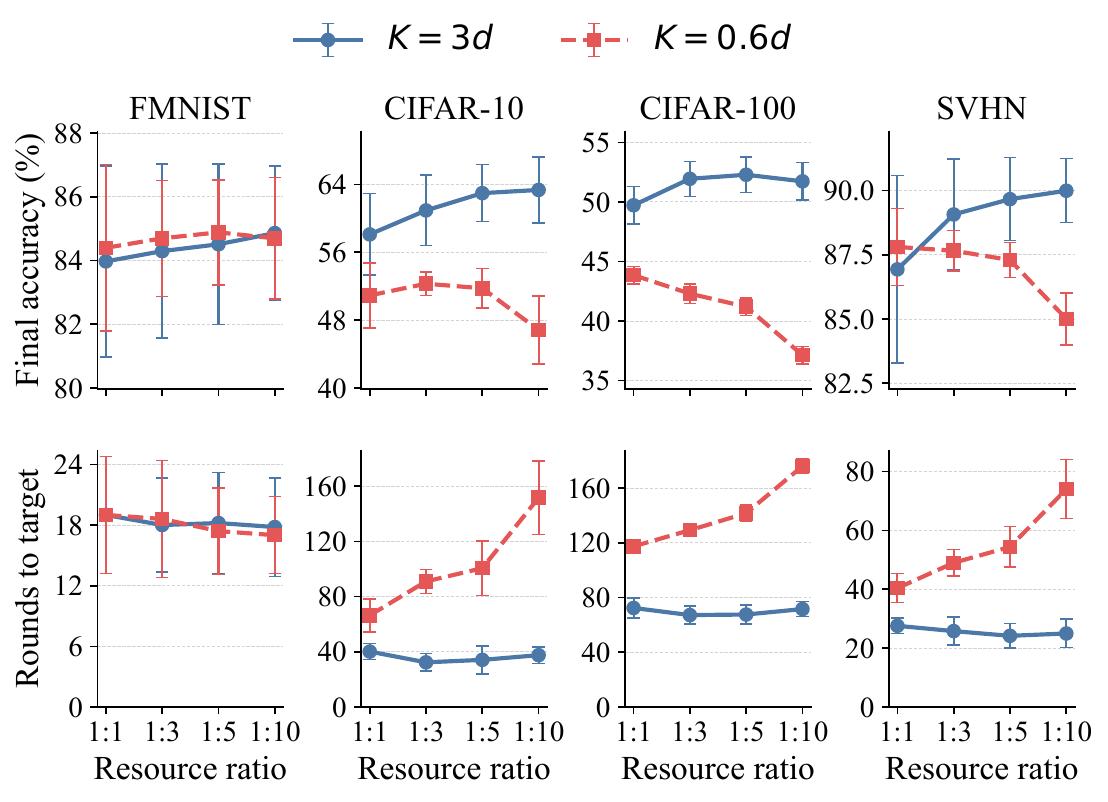}
\caption{Test accuracy and convergence rounds for $\mathrm{Dir}$($0.1$) across resource ratios and sparsification rates. Results are averaged over 5 seeds.
}
\label{fig:sparse}
\end{figure}

\section{Experimental Results}
\subsection{Experimental Setup}

\begin{table}[!t]
\centering
\caption{Matched uplink comparison with FedEns. FedEns and SHEFL retain
$0.3d$ and $0.1d$ coordinates per update, respectively. Each entry reports
accuracy (\%) / rounds to target.}
\label{tab:matched_uplink}
\footnotesize
\setlength{\tabcolsep}{2.5pt}
\renewcommand{\arraystretch}{1.08}

\begin{tabular}{@{}lcc@{}}
\toprule
Setting
& FedEns
& SHEFL \\
\midrule

C10 $\mathrm{Dir}(0.6)$
& {78.06$\pm$1.63} / 50.4$\pm$4.5
& \textbf{78.63}$\pm$0.90 / \textbf{45.8}$\pm$2.8 \\

C10 Path$(4)$
& 58.09$\pm$5.02 / 166.8$\pm$16.5
& \textbf{72.74}$\pm$2.76 / \textbf{88.2}$\pm$8.7 \\

C100 $\mathrm{Dir}(0.6)$
& 62.75$\pm$0.60 / 121.4$\pm$5.3
& \textbf{63.46}$\pm$0.48 / \textbf{108.6}$\pm$5.9 \\

C100 Path$(20)$
& 48.94$\pm$3.08 / 273.2$\pm$32.9
& \textbf{59.12}$\pm$0.95 / \textbf{148.2}$\pm$4.6 \\

\bottomrule
\end{tabular}

\vspace{0.5mm}
\end{table}
We evaluate SHEFL on FMNIST \cite{fmnist}, CIFAR-10/100
\cite{cifar10}, and SVHN \cite{svhn}. Data heterogeneity is
simulated using Dirichlet and pathological partitions. We use
$\mathrm{Dir}(0.6)$ for baseline comparisons and
$\mathrm{Dir}(0.1)$ for the resource allocation study. For the
pathological partition, each client receives 4 non-overlapping
shards \cite{mcmahan2017communication}, while 20 shards are
used for CIFAR-100. We employ a convolutional network with
2 convolutional layers and 2 hidden layers for FMNIST,
ResNet-18 for CIFAR-10 and SVHN, and ResNet-34 for
CIFAR-100 \cite{resnet}.

We compare SHEFL with FedAvg \cite{mcmahan2017communication},
FedProx \cite{fedprox}, FedNova \cite{fednova}, HeteroFL
\cite{heterofl}, FedBE \cite{fedbe}, and FedEns
\cite{shi2023fed}. For HeteroFL, HPCs and LPCs use width multipliers of 1.0 and
0.5. For the main comparison, FedAvg, FedProx, FedNova, and FedEns
apply TopK sparsification and retain $0.1d$ coordinates per client
update. FedBE transmits dense full model updates, while HeteroFL
transmits dense heterogeneous submodels. We use 100 clients with 50 HPCs and sample 5 HPCs and
5 LPCs per round. SHEFL uses $M=5$ global models.
Local training uses SGD with a learning rate of $10^{-2}$,
batch size 16, weight decay $10^{-3}$, and 5 local epochs. 
The learning rate is multiplied by 0.99 every 10 rounds.
We use $\mu=0.01$ for FedProx and follow
the original FedBE configuration with 5 samples. Communication rounds are set to 100 for FMNIST
and SVHN, 200 for CIFAR-10, and 400 for
CIFAR-100. The default SHEFL setting uses $r=M=5$, yielding
$k_H=k_L=0.1d$ and a total sparse uplink budget of $K=3d$.
The resource allocation study varies the $\mathrm{LPC}:\mathrm{HPC}$
ratio over $\{1:1,1:3,1:5,1:10\}$ under fixed total uplink budgets
$K\in\{3d,0.6d\}$. The tighter budget corresponds to
$k_H=k_L=0.02d$ when $r=M$.
 \subsection{Results}
 
Table~\ref{tab:acc_combined} compares final accuracy across data
heterogeneity settings under the default communication configurations
described above. SHEFL demonstrates comparable performance and improves over FedEns in all settings, with the largest gains appearing on pathological FMNIST and CIFAR-100.
The improvement is more pronounced when the direct ensemble baseline
degrades under fragmented client data, while the difference is small on
the easier Dirichlet FMNIST and SVHN settings. Other baselines achieve the highest accuracy in several settings, showing that the benefit of additional ensemble participation depends on the dataset and heterogeneity regime.

Figure \ref{fig:sparse} reports final accuracy and convergence rounds
(80\% for FMNIST and SVHN, 40\% for CIFAR-10, and 35\% for CIFAR-100)
under $\mathrm{Dir}(0.1)$ heterogeneity. With $K=3d$, increasing the HPC
allocation generally improves accuracy on CIFAR-10 and SVHN. Under the
tighter budget of $K=0.6d$, however, larger HPC allocations can reduce
accuracy and increase the rounds required to reach the target, indicating
that allocation proportional to computation is not consistently effective.
On CIFAR-10 with $K=0.6d$, the $1:1$ allocation reaches the target in
66.4 rounds, compared with 151.6 rounds for $1:10$. Under the same
per-round sparse budget, this corresponds to approximately $2.3\times$
fewer transmitted coordinates to reach the target.

Table~\ref{tab:matched_uplink} compares SHEFL against FedEns under
the same total sparse uplink budget per round. SHEFL reaches the
target accuracy (70\% for CIFAR-10, 50\% for CIFAR-100) in fewer rounds and achieves higher mean final accuracy in all four settings. The difference is largest under pathological partitions, where SHEFL improves final accuracy by 14.65 and 10.18 percentage points on CIFAR-10 and CIFAR-100, respectively.

\begin{figure}[!t]
\centering
\begin{minipage}[t]{0.66\columnwidth}
  \centering
  \includegraphics[width=\linewidth]{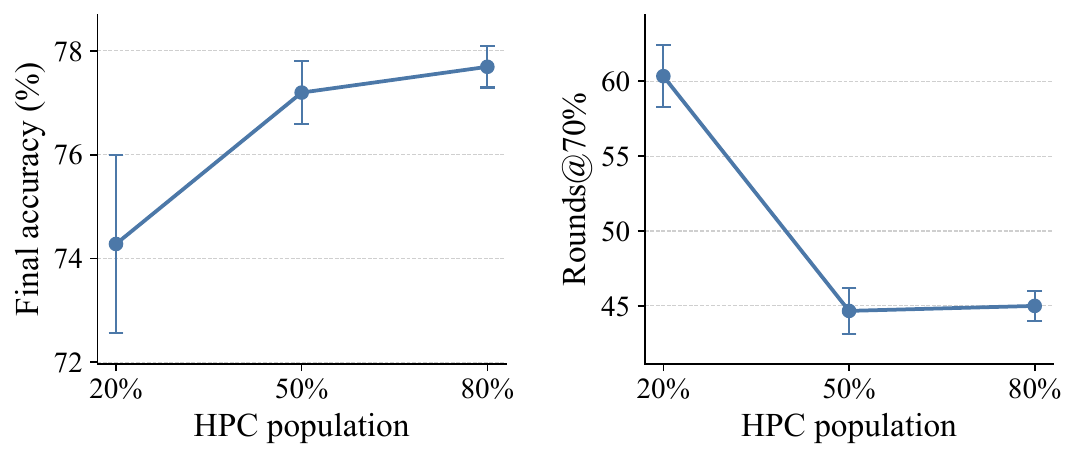}
  \vspace{1mm}
  {\footnotesize (a) HPC population.}
\end{minipage}
\begin{minipage}[t]{0.32\columnwidth}
  \centering
  \includegraphics[width=\linewidth]{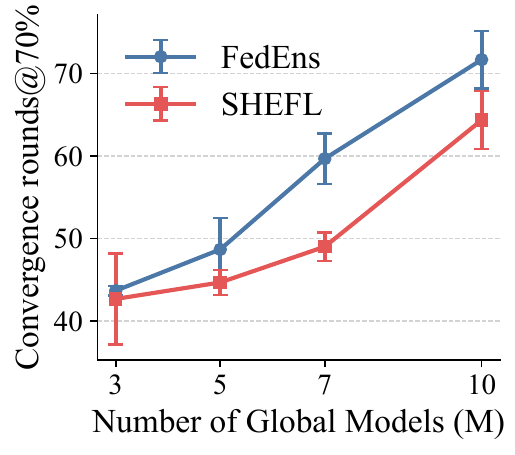}
  \vspace{1mm}
  {\footnotesize (b) Ensemble size.}
\end{minipage}
\caption{
(a) Test accuracy and convergence of SHEFL with varying numbers of HPCs. 
(b) Convergence rounds with increasing ensemble size for FedEns and SHEFL.
}
\label{fig:ablation_config}
\end{figure}



\begin{table}[!t]
\centering
\caption{Group-balanced aggregation under two resource allocations on
CIFAR-10 with $\mathrm{Dir}(0.6)$ and $M=5$. The total sparse uplink
budget is fixed at $K=3d$.}
\label{tab:agg_ablation}
\footnotesize
\setlength{\tabcolsep}{3pt}
\begin{tabular}{c cc cc}
\toprule
\multirow{2}{*}{Resource ratio}
& \multicolumn{2}{c}{Uniform}
& \multicolumn{2}{c}{Group-balanced} \\
\cmidrule(lr){2-3}\cmidrule(lr){4-5}
& Acc. & Rounds@70\%
& Acc. & Rounds@70\% \\
\midrule
$1:5$
& 76.22$\pm$0.55
& 50.4$\pm$5.2
& \textbf{78.63}$\pm$0.90
& \textbf{45.8}$\pm$2.8 \\

$1:1$
& 77.43$\pm$0.59
& 52.4$\pm$3.2
& \textbf{78.89}$\pm$1.01
& \textbf{46.4}$\pm$4.5 \\
\bottomrule
\end{tabular}
\end{table}
\subsection{Ablation studies}
Figure~\ref{fig:ablation_config}(a) examines the effect of the HPC
population. SHEFL reaches 70\% accuracy for all three tested HPC
fractions, including the 20\% HPC setting. Increasing the HPC fraction
from 20\% to 50\% reduces convergence rounds and improves accuracy, while a further
increase to 80\% gives incremental benefits. Figure~\ref{fig:ablation_config}(b) examines the effect of ensemble size under a fixed total sparse uplink budget of $K=3d$. Under the fixed budget $K=3d$, SHEFL reaches 70\% accuracy in fewer rounds than FedEns for all tested ensemble sizes, with larger differences at $M=7$ and $M=10$.

Finally, Table~\ref{tab:agg_ablation} verifies the effect of the proposed aggregation for ratios $1:5$ and $1:1$. Within each allocation, sparsification, local training, and client sampling are identical across the two aggregation rules. Group-balanced aggregation increases accuracy and reduces Rounds@70\% compared with uniform aggregation under both resource allocations, showing that 
communication budget allocation alone does not eliminate the
contribution imbalance caused by asymmetric ensemble
participation.
\section{Conclusion}
We propose SHEFL, an ensemble FL framework
that uses additional computation at HPCs under sparse uplink
constraints. SHEFL increases the number of contributors to each ensemble member through HPC participation, while controlling the resulting communication and contribution imbalance. Across four vision datasets, SHEFL achieves higher accuracy than FedEns in all evaluated heterogeneous settings and reaches the target accuracy faster under matched uplink budgets.

\bibliographystyle{IEEEbib}
\bibliography{strings,refs}

\end{document}